\pdfoutput=1
\documentclass[a4paper]{article}

\usepackage{INTERSPEECH2021}

\usepackage{hyperref}
\usepackage{cleveref}
\usepackage{subcaption}
\usepackage{url}
\usepackage{times}
\usepackage{latexsym}
\usepackage{amsfonts}
\usepackage{amsmath}
\usepackage{graphicx}

\usepackage[T1]{fontenc}
\usepackage[utf8]{inputenc}

\title{Team Hitachi @ AutoMin 2021: Reference-free Automatic Minuting Pipeline with Argument Structure Construction over Topic-based Summarization}

\name{Atsuki Yamaguchi$^{\text{*}}$, Gaku Morio$^{\text{*}}$, Hiroaki Ozaki$^{\text{*}}$, Ken-ichi Yokote and Kenji Nagamatsu}
\address{
Research and Development Group\\ 
Hitachi, Ltd. \\
Kokubunji, Tokyo, Japan
}
\email{\{atsuki.yamaguchi.xn, gaku.morio.vn, hiroaki.ozaki.yu, kenichi.yokote.fb, kenji.nagamatsu.dm\}@hitachi.com}

\begin{document}

\maketitle
\begin{abstract}
This paper introduces the proposed automatic minuting system of the Hitachi team for the First Shared Task on Automatic Minuting (AutoMin-2021).
We utilize a reference-free approach (i.e., without using training minutes) for automatic minuting (Task A), which first splits a transcript into blocks on the basis of topics and subsequently summarizes those blocks with a pre-trained BART model fine-tuned on a summarization corpus of chat dialogue.
In addition, we apply a technique of argument mining to the generated minutes, reorganizing them in a well-structured and coherent way.
We utilize multiple relevance scores to determine whether or not a minute is derived from the same meeting when either a transcript or another minute is given (Task B and C).
On top of those scores, we train a conventional machine learning model to bind them and to make final decisions.
Consequently, our approach for Task A achieve the best adequacy score among all submissions and close performance to the best system in terms of grammatical correctness and fluency. For Task B and C, the proposed model successfully outperformed a majority vote baseline.

\end{abstract}
\noindent\textbf{Index Terms}: automatic minuting, summarization, argument mining, topic segmentation

{
\let\thefootnote\relax\footnotetext{* Equal contribution}
}

\section{Introduction}
\label{intro}
This paper introduces the proposed automatic minuting system of the Hitachi team for the First Shared Task on Automatic Minuting (AutoMin-2021) \cite{Ghosal2021}.
The shared task consists of a main task (Task A) and two subtasks (Task B and C).
Task A aims at generating minutes from meeting transcripts. 
Task B involves identifying whether or not a given minute belongs to a given transcript.
Similar to Task B, Task C requires a model to identify whether or not two given minutes belong to the same or different meeting(s). 

Our initial approach for Task A was to build a deep reinforcement learning-based end-to-end minuting system \cite{chen-bansal-2018-fast, Zou_Zhao_Kang_Lin_Peng_Jiang_Sun_Zhang_Huang_Liu_2021} using the provided reference minutes.
The system consists of a pick-up module and a summarization module.
The former selects important utterances from transcripts. We also expected that this module could be useful for Task B and C to measure the relevance.
The latter summarizes utterances that are selected by the pick-up module.
Unfortunately, this approach did not perform well in our preliminary experiments. Through an error analysis, we found that (1) sentences in minutes did not always have their corresponding utterances in transcripts; (2) sentences in minutes highly depended on each annotator's aspect, and (3) reference resolution (resolving who ``I'' is or who``you'' is, etc.) was necessary because minutes are often written from a third-person perspective, while the transcripts are often from a first-person perspective.
These (1) and (2) observations also made solving Task B and C difficult.
Consequently, we figured out that these ``ground-truth'' minutes did not contribute to summarizing transcripts.

Hence, we subsequently adopt a reference-free approach for Task A and did not utilize reference minutes to generate summaries given that it is challenging for a machine learning model to learn the relationship between a summary and a transcript due to (1) and (2). For Task B and C, we employ a conventional feature-based machine learning method to make classifications.

Our pipeline for Task A mainly consists of segmentation, summarization, and argument mining modules.
The segmentation module is designed to extract utterances by \textit{block}, the utterances of which should mention the same topic in a transcript. At the same time, this module can filter out irrelevant utterances by excluding them from a block. We build a Longformer-based segmentation module trained with our manually annotated data, which are sampled from English training data. Surprisingly, the topic boundaries are well matched between our two annotators with an average agreement ratio of 0.818.
Given an utterance block, the summarization module generates its summary. To achieve this, we utilize a pre-trained BART model fine-tuned on the SAMSum corpus \cite{gliwa-etal-2019-samsum}. 
Because the model can partially solve reference problem (3) through the segmentation and summarization modules, we can mostly have a comprehensive summarization of an input transcript.
Finally, we structure the summarized blocks and formulate a resulting minute with an off-the-shelf argument mining parser \cite{morio-etal-2020-towards}.
This module is derived from our observation that most of the reference minutes are structured by itemization. 
Although a lot of variety exists in the formats, aspects, and perspectives of the structures mentioned in (2), 
we hypothesize that any coherent structures would improve readability and enable readers to recognize certain aspects and perspectives of the minutes.
To compose such structures in a minute, we use the argument mining parser, which can predict argumentative labels and reasons for each sentence.
In addition, our system can also handle transcripts in Czech. We simply add an mBART \cite{liu-etal-2020-multilingual-denoising} translation module that converts an input transcript from Czech into English and also changes its output minute from English to Czech.

Our experimental results on the English test set show that our automatic minuting system achieves the best adequacy score among all submissions, indicating that it can generate a minute with major topics covered in each meeting. In addition, our system exhibits close scores to the best system in terms of grammatical correctness and fluency.

For Task B and C, we define surface-oriented features rather than semantics-oriented ones.
We first define relevance scores, such as tf-idf \cite{tfidf} cosine similarity, named entity overlap ratio, date consistency, and BERTScore \cite{bert-score}. On top of the relevance scores, we trained machine learning models, such as support vector machine (SVM) \cite{svm}, logistic regression \cite{logistic_regression}, and random forest \cite{random_forest}. We pick up the best performing model from them to make a final classification. The results demonstrate that our approach outperforms a majority vote baseline, showing its effectiveness in classifying minutes.

\section{Task A: Automatic Minuting}

\subsection{Overview}
\begin{figure*}[t]
    \centering
    \includegraphics[width=0.8\linewidth]{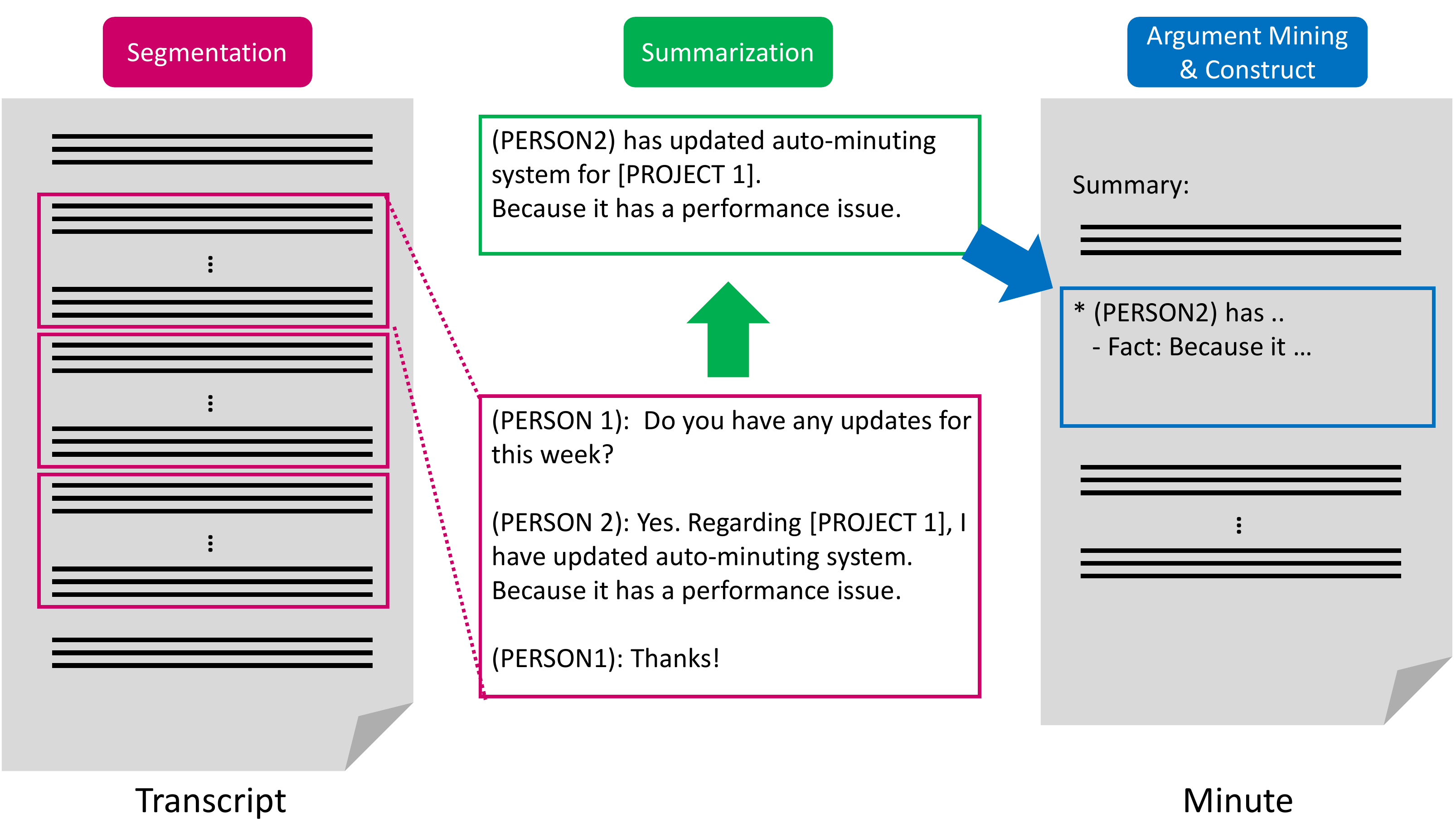}
    \caption{Overview of our approach. We first split a transcripts into blocks and then summarize them using a pre-trained model. Finally, we compose all summaries into minutes with a technique of argument mining.}
    \label{fig:overview}
\end{figure*}

We propose a module-based approach to generate minutes automatically from transcripts. 
\Cref{fig:overview} shows an overview of our approach. It mainly consists of the following three modules: segmentation, summarization, and argument mining. We first split an input transcript into topic-specific blocks using a Longformer-based model \cite{beltagy-etal-2020-longformer}. Then we summarize texts in each block with BART~\cite{lewis-etal-2020-bart} fine-tuned on the SAMSum \cite{gliwa-etal-2019-samsum}, a summarization corpus of chat dialogue.
Finally, we structure the summarized blocks and formulate a resulting minute with an argument mining parser. Note that our approach is applicable to transcripts written in both English and Czech by utilizing a pre-trained mBART model.

\subsection{Segmentation Module}
Given that state-of-the-art neural-network-based summarization models cannot process a long transcript at a time, it must be split into some segments so that the models can accommodate them. Our method segments such transcripts into blocks by topic, allowing us to generate a summary per topic in the later stage.

\subsubsection{Approach}
We utilize a sequence labeling task (BIO-tagging) to split a transcript into topic-specific blocks and employ a pre-trained Longformer\textsubscript{LARGE} model followed by a bidirectional Long Short-Term Memory (LSTM) \cite{hochreiter-1997-lstm} to solve the task. For each transcript, we first split it into sentences while inserting a special \texttt{[SEP]} token at the beginning of each sentence, followed by tokenization with a pre-trained Longformer tokenizer. Then we feed the tokenized transcript into the Longformer+LSTM model and predict a BIO tag for each \texttt{[SEP]} token. The motivation behind the use of Longformer and LSTM is that (i) Longformer can handle long sequences unlike other conventional Transformer-based models \cite{vaswani-etal-2017-attention}; (ii) LSTM can force the model to learn the sequential nature of the task that each label is dependent on its adjacent labels.

More formally, given a transcript $\mathcal{T}$ composed of $n$ sentences and its corresponding topic BIO labels, we first split $\mathcal{T}$ into sentences $S_\mathcal{T}=\{s_1, \dots, s_n\}$, while adding a \texttt{[SEP]} token at the beginning of each sentence. We then tokenize them using a pre-trained Longformer tokenizer and obtain their token-level representation $\mathbf{X} = (\mathbf{x}_1, \dots, \mathbf{x}_m)$. We feed $\mathbf{X}$ into the Longformer\textsubscript{LARGE} model and pick up its output representation of the \texttt{[SEP]} tokens $\mathbf{H}_\mathcal{T}=(\mathbf{h}_1, \dots, \mathbf{h}_n), \mathbf{h}_i \in \mathbb{R}^{D}$, where $D$ corresponds to the dimensionality of each hidden layer in Longformer. $\mathbf{H}_\mathcal{T}$ is subsequently passed to a one-layer bidirectional LSTM, which yields $\mathbf{H}_\mathcal{T}'=(\mathbf{h}_1', \dots, \mathbf{h}_n'), \mathbf{h}_i' \in \mathbb{R}^{2 \times D}$. Finally, we put $\mathbf{H}_\mathcal{T}'$ into a final output linear layer and obtain $\mathbf{O} = (\mathbf{o}_1, \dots, \mathbf{o}_n)$, where $\mathbf{o}_i \in \mathbb{R}^3$ is the $i$-th output of $S_\mathcal{T}$. The task is trained with the token-level cross-entropy loss averaged over the $\texttt{[SEP]}$ tokens:
\begin{equation}
    \mathcal{L} = -\frac{1}{n}\sum_{i=1}^{n}\sum_{j=1}^{3} y_{ij} \log p_{ij}(s_i),  \nonumber
\end{equation}
\noindent where $p_{ij}(s_i)$ represents the probability of the $i$-th input sentence $s_i$ predicted as \textit{beginning of a topic} ($j=1$), \textit{inside a topic} ($j=2$), or \textit{outside of a topic} ($j=3$) by the Longformer+LSTM model. $y_{ij}$ is the corresponding target label.

\subsubsection{Topic Boundary Annotation}
We manually annotated ten different transcripts in English training data, of which we utilized nine transcripts for training and one for validation. We assigned topic boundary labels per sentence because the Longformer+LSTM model predicts topic BIO labels by sentence.
We defined the average agreement ratio between annotators as follows:
Let $\mathcal{B}_1: \left\{b_{1,1}, \dots \right\}$ be a set of all blocks given by annotator 1
and $\mathcal{B}_2:  \left\{b_{2,1}, \dots \right\}$ be a set of all blocks given by annotator 2,
where $b$ represents a block $[s_i, \dots, s_j]$.
Then the agreement rate ($a_{1,2}$) is given by 
\begin{equation}
    a_{1,2} = \frac{1}{|\mathcal{B}_1|} \sum_{b_1 \in \mathcal{B}_1} \underset{b_2 \in \mathcal{B}_2}{\max} 
    \frac{|b_1 \cap b_2|}{|b_1|}. \nonumber
\end{equation}
The average of $a_{1,2}$ and $a_{2,1}$ for the validation transcript\footnote{We used \texttt{en\_train\_009.txt} as our validation transcript.} was 0.818. 

\subsubsection{Implementation Details}
We implemented the segmentation model using PyTorch \cite{paszke-etal-2019-pytorch} and the \texttt{transformers} library \cite{wolf-etal-2020-transformers}.
We fine-tuned the model for 100 epochs with a batch size of 1 and utilized the best weights for inference that achieved a validation accuracy of 0.850 at 510 steps. We set the learning rate to $10^{-5}$ and utilized the Adam optimizer \cite{adam} with a linear warmup of the first 5\% of steps and a gradient accumulation every five steps.
As pre-processing, if the total number of tokens in $\mathcal{T}$ exceeded the maximum length of input tokens for Longformer $L_{max}=4096$, we created chunks that can hold up to $L_{max}$ tokens and put sentences in each chunk with a stride of 1024 tokens. We then fed each chunk into the segmentation model and obtained its predicted BIO labels. To concatenate predicted labels from multiple chunks, we prioritized the predicted label(s) of a prior chunk whenever prediction results were duplicated between the two chunks.

\subsection{Summarization Module}
The summarization module generates a summary for each block with a pre-trained language model. Specifically, we use a pre-trained BART model fine-tuned on CNNDailyMail \cite{nallapati-etal-2016-abstractive,cnndailymail} and the SAMSum corpus, which is a large corpus of chat dialogue consisting of over 16k samples with abstractive summaries. With the pre-trained model, we can generate a concise and fluent summary given a dialogue block.
Because each summary in the SAMSum corpus extracts essential information in its dialogue, is written in the third person, and includes the names of speakers, we expect that a summarization model trained with the corpus should be able to resolve references.

\subsubsection{Implementation Details}
We employed a pre-trained BART\textsubscript{LARGE} model fine-tuned on the CNNDailyMail and SAMSum datasets as our summarization model.\footnote{Namely \texttt{philschmid/bart-large-cnn-samsum} available at \url{https://huggingface.co/philschmid/bart-large-cnn-samsum}} The weights were downloaded via the \texttt{transformers} library.
For pre-processing, we added a speaker tag, such as ``PERSON1:'', to the beginning of each utterance as a prefix. If the number of tokens in a block exceeded the maximum length for BART $L_{max}=1024$, we truncated some utterances in the block in advance to avoid generating incomplete sentences.
During post-processing, we replaced some misspelled or unnatural tokens generated by the model with correct ones in accordance with pre-defined regular expression rules to make the generated summary more consistent.

\begin{figure*}[t]
    \centering
    \includegraphics[width=0.7\linewidth]{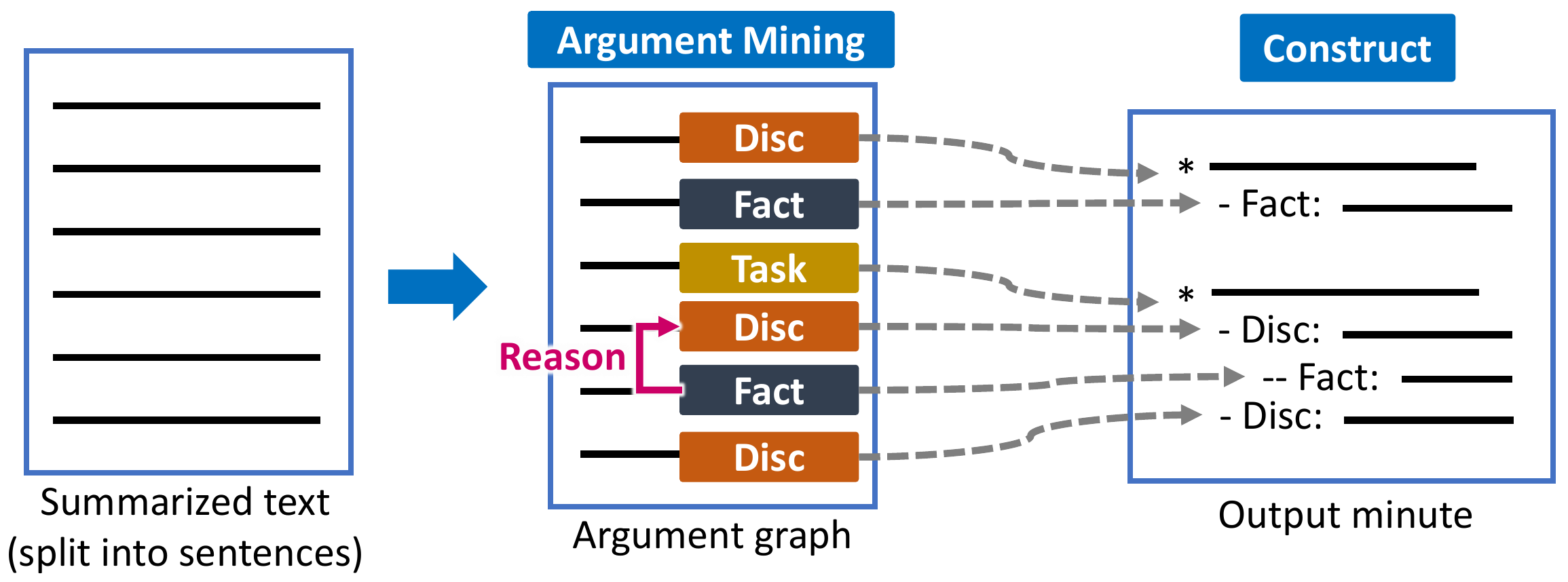}
    \caption{Overview of the argument mining module. The summarized text in a block is split into sentences. The argument mining parser predicts an argument graph for the sentences. Finally, we structure a minute using itemization to make it coherent and to improve readability. Note that the first sentence in the argument graph and Task sentences should be the root for the itemization. We do not change the order of sentences to prevent corrupting the fluent and coherent summarized text.}
    \label{fig:argmin}
\end{figure*}

\begin{table*}[t]
    \centering
    \begin{tabular}{llllllll}
        \hline
         & Language & Adequacy & Grammatical correctness & Fluency & ROUGE 1 & ROUGE 2	& ROUGE L \\
         \hline
         Ours & en & \textbf{4.25} & 4.34 & 3.93 & 0.217 & 0.0595 & 0.116 \\
         Average of all teams &  & 2.81 & 3.25 & 2.92 & 0.203 & 0.0458 & 0.114 \\
         Best in all teams &  & 4.25 & \textbf{4.45} & \textbf{4.27} & \textbf{0.282} & \textbf{0.0655} & \textbf{0.159} \\
         \hline
         Ours & cs & 2.69 & 1.25 & 2.06 & 0.159 & 0.0251 & 0.0694 \\
         \hline
    \end{tabular}
    \caption{Task A results on the test set. The adequacy, grammatical correctness and fluency are based on a manual evaluation by two annotators and assessed on a Likert Scale of 1 to 5, where 1 represents the worst and 5 refers to the best. All scores the are averaged across all test set samples and based on official results provided by the organizers.}
    \label{tab:task-A-result}
\end{table*}

\subsection{Argument Mining Module}

This module creates a structured minute from summarized texts.
We use a concept of argument mining \cite{lawrence-reed-2019-argument}, an attractive research area that focuses on mining arguments such as reasons behind a claim.
We provide an off-the-shelf argument mining parser \cite{morio-etal-2020-towards} to construct structures.
The parser predicts an argument graph based on a concept of the Cornell eRulemaking Corpus (CDCP) \cite{park-cardie-2018-corpus}.
With the concept of the CDCP, we can predict an argumentative label for each proposition and relationship between propositions.
In our system, we split the summarized text (i.e., a result from the summarization module in a block) into sentences by spaCy \cite{spacy}, recognizing each sentence as a proposition.

\subsubsection{Argument Graph}
We predict an argumentative label for each proposition using the argument mining parser.
There are three types of proposition labels having the following description.
\begin{enumerate}
  \item \textbf{Task} is a subjective sentence that contains a specific policy (e.g., what should be done). We expect that sentences with this label should include an important topic or discussion aspect. The original label name in CDCP is Policy.
  \item \textbf{Fact} is an objective sentence. We map Testimony and Fact labels that are originally introduced in CDCP into this label.
  \item \textbf{Disc} is a subjective sentence such as an opinion and claim. The original label name in CDCP is Value.
\end{enumerate}
In addition to label each proposition, we predict argumentative relationships between propositions.
The argument mining parser predicts a Reason or Evidence relationship, which is a support relationship between the propositions.
We obtain an argument graph by combining both the proposition labels and relationships.

\subsubsection{Structured Minute Construction}
Our approach to construct the structured minutes (shown in \Cref{fig:argmin}) from the predicted argument graph is as follows:

\begin{enumerate}
  \item We assume that the first sentence in the summarized text and sentences that have a Task label are the \textit{root} item of the structure. We expect these sentences represent a specific topic or aspect of the summarization. The root items are represented as ``* ...'' as shown on the right in \Cref{fig:argmin}.
  \item Subsequent sentences for the preceded root item become a child item. For instance, we represent a child item with a predicted proposition label of Fact as ``- Fact: ...''.
  \item A sentence that has an outgoing relationship to a preceded sentence becomes a child item for the preceding sentence. We expect that we can make a supplemental or supporting sentence for its previous sentence clear for readers. For example, in \Cref{fig:argmin}, the Fact sentence has an outgoing Reason relationship in the Disc sentence. We represent this by ``- - Fact: ...''.
\end{enumerate}

\subsection{Results}
\Cref{tab:task-A-result} shows our submission results of the auto minuting task on the test set.
First, we observe that our English system achieved the best adequacy score of 4.25 out of 5 among all submissions, suggesting that our approach can adequately capture major topics appearing in each meeting transcript. 
Second, our approach also exhibited similar human evaluation scores to the best system by 0.11 for grammatical correctness and 0.34 for fluency and far better scores than the average ones. These results demonstrate that our automatically generated minutes are generally grammatically consistent and have some fluency and coherency.
Finally, our ROUGE scores were narrowly better than the average scores, and huge gaps existed between our results and the best ones in terms of the automatic evaluation metrics. These results show a different trend compared to the ones involving the human evaluation.

We further investigate the relationship between human and automatic evaluations.
\Cref{tab:corr} shows the correlation coefficients between the human evaluation metrics and the automatic ones.
We used average scores for the automatic metrics. Each metric has a weak correlation ranging from 0.42 to 0.544.
Instead of averaging the automatic metrics over reference minutes, we took the maximum values among the reference minutes in \Cref{tab:corr-max}. We only used samples with more than two reference minutes.
Compared to \Cref{tab:corr}, \Cref{tab:corr-max} shows a strong correlation between adequacy and ROUGE 1.
In contrast, grammatical correctness and fluency are less correlated with the automatic metrics, especially on ours.
We discuss the relationship between the human evaluation metrics and the automatic ones in \S\ref{task-A-discussion}.

For the Czech results, our translation-based approach appeared to have failed to generate natural minutes because all of our Czech scores were far lower than those in English.

\subsection{Discussion} \label{task-A-discussion}
\subsubsection{Automatic vs. Human Evaluations}
\begin{table}[t]
    \centering
    \begin{tabular}{llll}
        \hline
         &  ROUGE 1 & ROUGE 2 & ROUGE L \\
         \hline
         Adequacy & 0.544 & 0.496 & 0.480 \\
         Correctness & 0.430 & 0.433 & 0.451 \\
         Fluency & 0.494 & 0.420 & 0.483 \\
         \hline
    \end{tabular}
    \caption{Correlation coefficients between human and automatic evaluations.}
    \label{tab:corr}
\end{table}

\begin{figure*}[t]
    \centering
    \begin{minipage}[b]{0.45\hsize}
        \centering
        \includegraphics[width=0.8\textwidth]{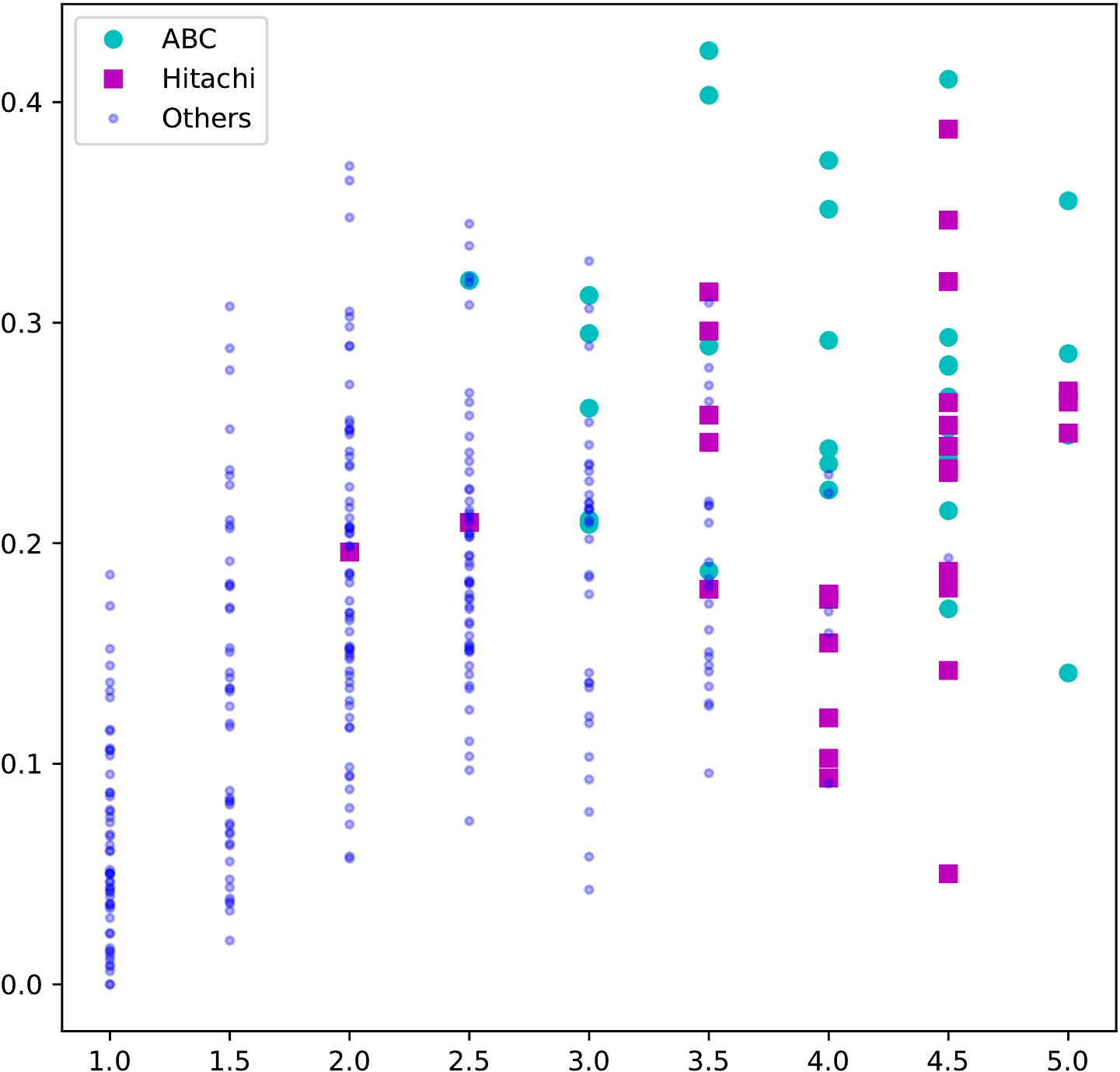}
        \subcaption{Average}
    \end{minipage}
    \begin{minipage}[b]{0.45\hsize}
        \centering
        \includegraphics[width=0.8\textwidth]{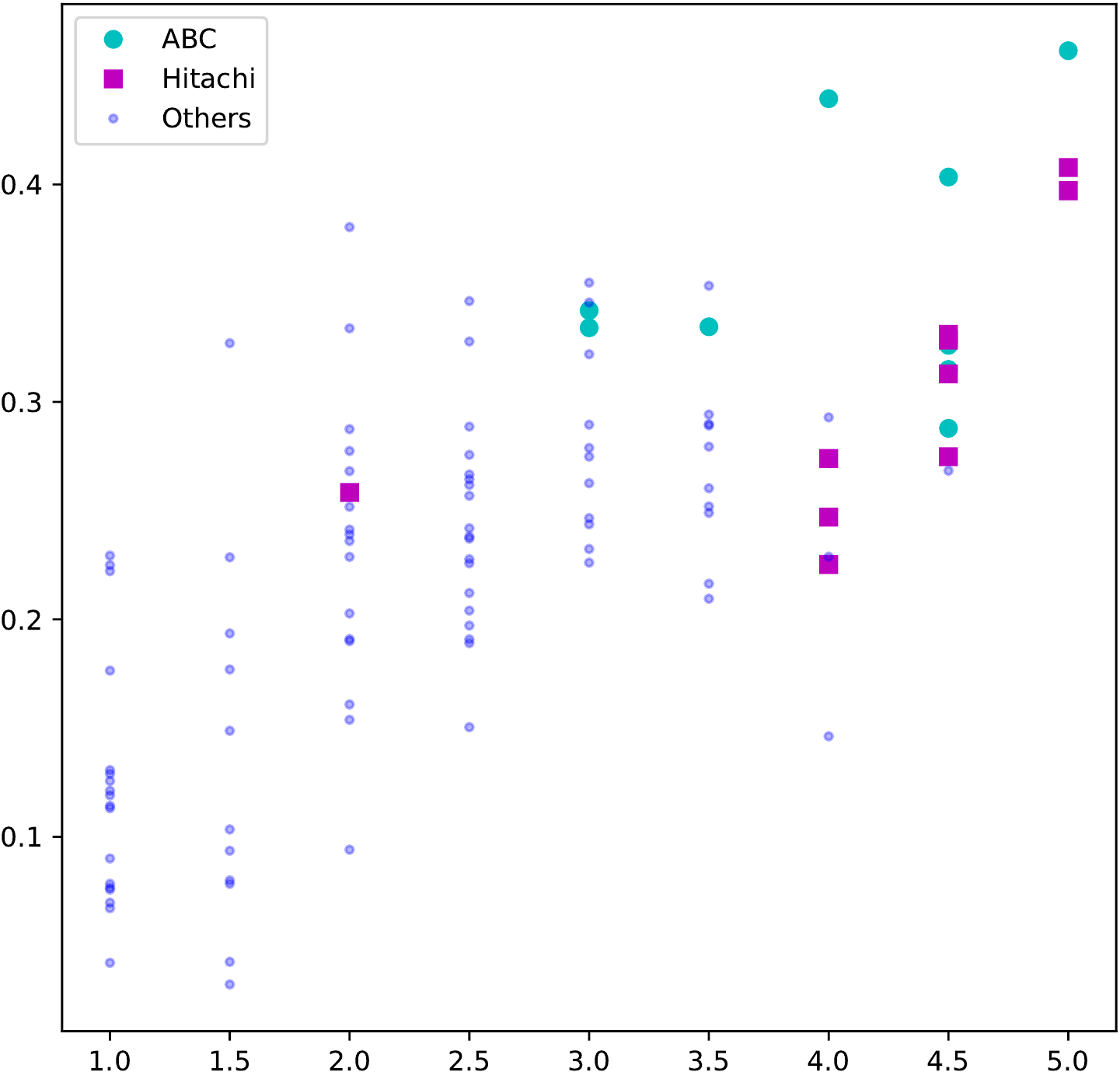}
        \subcaption{Max Pooling for more than two reference minutes}
    \end{minipage}
    \caption{Adequacy vs. ROUGE 1 scores. The horizontal axis represents Adequacy of each generated minute for the test set. The vertical axis represents ROUGE 1 scores.}
    \label{fig:corr}
\end{figure*}

\begin{table}[t]
    \centering
    \tabcolsep 6pt
    \begin{tabular}{lllll}
        \hline
         &  Data & ROUGE 1 & ROUGE 2 & ROUGE L \\
         \hline
         Adequacy & All & 0.715 & 0.580 & 0.549 \\
         & Ours & 0.940 & 0.575 & 0.804 \\
         Correctness & All & 0.408 & 0.333 & 0.365 \\
         & Ours & 0.188 & -0.218 & -0.122 \\
         Fluency & All & 0.547 & 0.397 & 0.486 \\
         & Ours & 0.195 & -0.0132 & 0.358 \\
         \hline
    \end{tabular}
    \caption{Correlation coefficients between human and automatic evaluations. Instead of averaging automatic metrics over reference minutes, we take maximum values among the reference minutes. We only use samples that have more than two reference minutes. ``All'' indicates that we include all of the participants' data, and ``Ours'' means only our results were used. }
    \label{tab:corr-max}
\end{table}

The comparison between \Cref{tab:corr}, \ref{tab:corr-max} and \Cref{fig:corr} indicates more than two types of adequate minutes in general, and those types are not similar to each other. This supports our hypothesis that the minutes are varied because of each annotator's diverse aspects as mentioned in \S\ref{intro}.
In addition, the results indicate that adequacy can be evaluated using word overlap between a target minute and the closest reference minute (\Cref{tab:corr-max}). This is consistent with the results of Task C, which we describe in the latter section.

Grammatical correctness and fluency were not related to any automatic metrics for ours shown in \Cref{tab:corr-max}, even though our model achieved high performance in the human evaluations.
This could be a nature of our reference-free approach. 
If a few variations did not occur in adequate minutes as we mentioned in \S\ref{intro},
the generated minutes using the reference-free approach may have grammatically correct and adequate sentences. However, those sentences are very different from the reference ones. In this case, measuring the grammatical correctness and fluency using N-gram based metrics will be difficult. 

The high human evaluation scores of grammatical correctness and fluency indicate a merit of our reference-free approach, which already captured grammatical correctness and fluency through pre-training on massive corpora. Compared with training an end-to-end model from scratch, which may require a lot of effort to capture grammatical correctness and fluency, we do not need any exhausting training.

\subsubsection{Translation Strategy}
In our preliminary experiments, we translated generated English minutes into Japanese (our mother tongue) and found they were acceptable. However, our Czech system did not perform well on the Czech test set (\Cref{tab:task-A-result}). 
According to Liu et al. \cite{liu-etal-2020-multilingual-denoising}, the translation accuracy of Japanese and Czech is almost the same. 
Low quality in transcript translations might lead to low scores on the Czech data, because converting transcripts in spoken language is considered to be more difficult than converting documents in written language. 

\begin{figure}[t]
    \centering
    \includegraphics[width=0.8\linewidth]{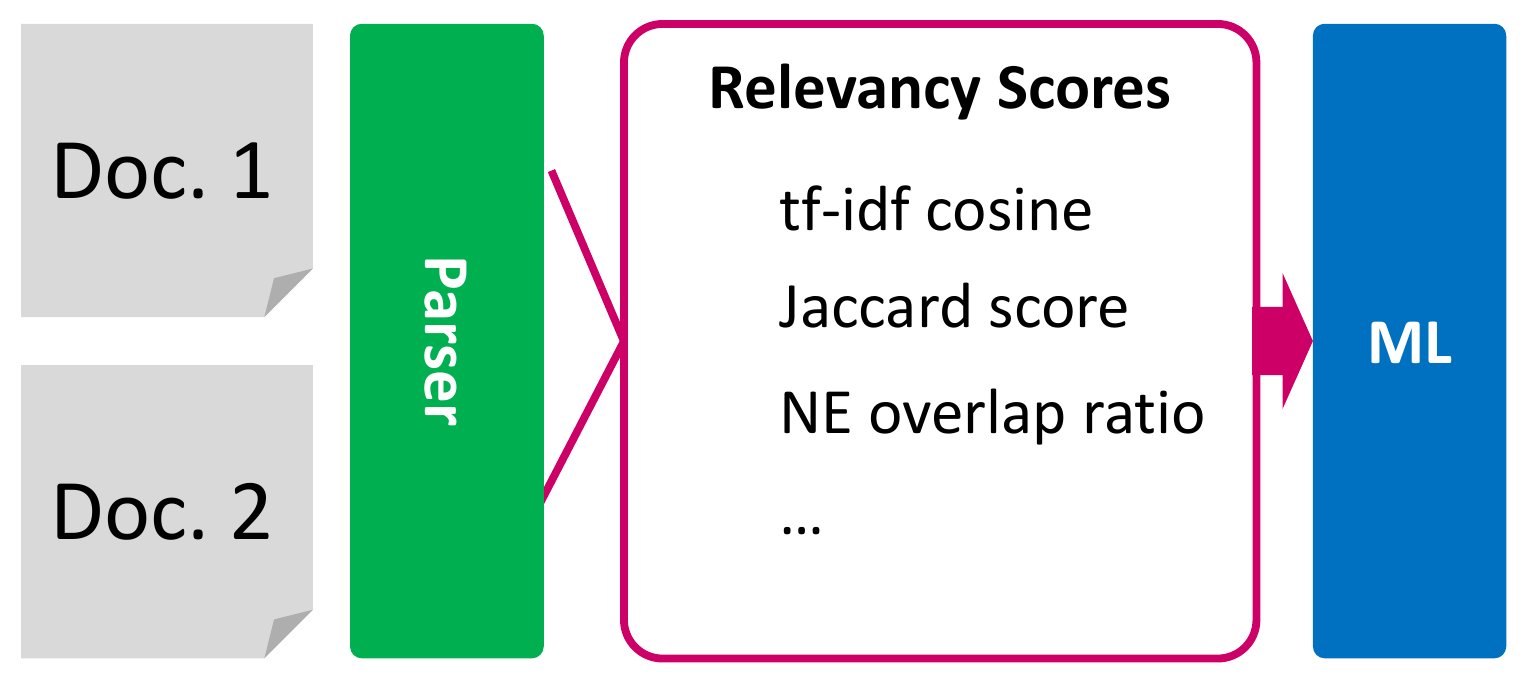}
    \caption{Overview of our approach for Task B and C. We first compute relevancy scores when two documents are given, then train a conventional machine learning model on top of the scores.}
    \label{fig:taskbc}
\end{figure}

\begin{table*}[t!]
    \centering
    \begin{tabular}{lll}
    \hline
        Action & Argument & Detail \\
        \hline
        LABEL & Label name (topic, title, date, etc.) & Set label to the $b_0$ and discard $b_0$.\\
        ADD & - & Add a parent-child relationship between $s_0$ and $b_0$ and add $b_0$ to the stack. \\
        REPLACE & - &  Add a parent-child relationship between $s_1$ and $b_0$ and add $b_0$ to the stack. \\
        ARC & - & Add a parent-child relationship between $s_0$ and $b_0$ and discard $b_0$. \\
    \hline
    \end{tabular}
    \caption{Action set of our transition-based minute parser.}
    \label{tab:transition}
\end{table*}

\section{Task B and C: Pair Classification}

\subsection{Overview}

\Cref{fig:taskbc} shows an overview of our approach.
We first define relevancy scores such as tf-idf \cite{tfidf} cosine similarity, named entity overlap ratio, date consistency. and BERTScore. On top of the relevancy scores, we trained machine learning models such as SVM \cite{svm}, logistic regression \cite{logistic_regression}, random forest \cite{random_forest} and multi-layer perceptron (MLP). We picked up the best performing model from them.

\subsection{Pre-processing}

To extract each relevancy score, we first applied a minute parser that can analyze a structure of lines in a minute.
The minute parser can find mainly the title, date, attendees, topic(s), subtopics and item lines with their parents or children.
We employed a transition-based parsing system \cite{yamada-matsumoto-2003-statistical, nivre-etal-2006-labeled} as our minute parser.
A transition-based parser usually consists of a stack and a buffer that holds unprocessed data.
Then, a certain action will be taken in accordance with the contents in the stack and the buffer.
We used random forest for the action predictor and trained it on manually annotated data sampled from the English training data.
\Cref{tab:transition} lists actions that we defined. Here, $b_0$ is the top element of the buffer, $s_0$ is the top element of the stack, and $s_1$ is the second top element in the stack.

\subsection{Features}

\subsubsection{tf-idf Score}
We computed a cosine similarity of tf-idf \cite{tfidf}. 
Let $\mathcal{D}_1$ be an input document. For Task B,  $\mathcal{D}_1$ is a transcript.
Let $\mathcal{D}_2$ be another input document. For both Task B and C cases, $\mathcal{D}_2$ is a minute.
We applied some prepossessing steps such as stop word removal using NLTK\footnote{\url{https://www.nltk.org/}}, punctuation mark normalization, and exclusion of words with fewer than three characters or with more than fourteen characters. 

\subsubsection{Jaccard Similarity Coefficient}
We computed the Jaccard similarity coefficient \cite{jaccard_similarity} of vocabularies in $\mathcal{D}_1$ and $\mathcal{D}_2$. We excluded terms in the NLTK stopword list and terms fewer than three characters or more than fourteen characters.

\subsubsection{NE Overlap}
Here we define a named entity as an anonymized token such as ``[PERSON X],'' ``[PROJECT X],'' and ``[ORGANIZATION X].''
We extracted them with regular expressions.
Let $\mathcal{E}$ be a function that extracts a set of all named entities in a given document. Then, the NE overlap score is calculated as
\begin{equation}
    \text{NE Overlap} := \frac{\left|\mathcal{E}(\mathcal{D}_1) \cap \mathcal{E}(\mathcal{D}_2) \right|}{\left|\mathcal{E}(\mathcal{D}_1) \cup \mathcal{E}(\mathcal{D}_2) \right|} . \nonumber
\end{equation}
When either $\mathcal{E}(\mathcal{D}_1)$ or $\mathcal{E}(\mathcal{D}_2)$ is $\emptyset$, we define NE overlap as zero.

\subsubsection{Date Consistency}
We extracted dates with a Python \texttt{dateutil.parser}\footnote{\url{https://dateutil.readthedocs.io/en/stable/parser.html}} from a ``date line.'' We defined four dimensional features, and then those of dimensions representing whether or not each year, month, day, and hour was consistent between $\mathcal{D}_1$ and $\mathcal{D}_2$.

\subsubsection{BERT Score}
We use BERTScore \cite{bert-score} to compute the semantic similarity between documents.
Because the transcript and minute documents are much longer for BERTScore, we need to split them.
Here we split $\mathcal{D}_1$ and $\mathcal{D}_2$ into $N$ chunks, i.e., $(\mathcal{D}_{1,1}, ... \mathcal{D}_{1,N})$ and $(\mathcal{D}_{2,1}, ... \mathcal{D}_{2,N})$.
We compute the BERTScore for each chunk pair, i.e., $s(i) = \textsc{BERTScore} (\mathcal{D}_{1,i}, \mathcal{D}_{2,i})$.
The averaged score $\frac{1}{N} \sum_{i=1}^{N} s(i)$ serves as the feature value.
In this study, we selected $N=4$.

\subsection{Model Selection}

We tuned the models with a hyperparameter optimization framework named Optuna \cite{Akiba:2019}\footnote{\url{https://www.preferred.jp/en/projects/optuna/}}.
Through the optimization, we applied 10-fold cross validation to avoid overfitting. To evaluate the validation score, we merged official dev and fold-out sets.
The best hyperparameters for each task are shown in \Cref{tab:task-bc-param}.
Finally, we used an ensemble average of the 10 cross-validation models to predict outputs of the test set.  

\begin{table}[t!]
    \centering
    \tabcolsep 5pt
    \begin{tabular}{llll}
        \hline
         & & Model & Hyperparameter values \\
         \hline
         Task B & en & SVM & kernel: linear, C: 8.05, $\gamma$: 0.00815 \\
         & cs & SVM & kernel: linear, C: 9.81, $\gamma$: 0.00172 \\
         \hline
         Task C & en & MLP & hidden dim: 64, layer: 3\\
         & cs & SVM &  kernel: rbf, C: 2.87, $\gamma$: 2.11 \\
         \hline
    \end{tabular}
    \caption{Hyperparameters for the best models of our Task B and C submissions.}
    \label{tab:task-bc-param}
\end{table}

\subsection{Results}
\Cref{tab:task-bc-result} shows the evaluation results on the test dataset.
We compare them with majority vote baselines, which always give FALSE for all data points.
All our submitted models outperformed the baselines with high F1 scores ranging from 0.659 to 0.904.
The task C English model only had a slightly better result than the baseline. This may have been due to the MLP model because it was prone to overfitting our hand-crafted features.

In our preliminary experiments,  features based on term overlaps (i.e., tf-idf and NE) contributed to the models' performance well.
This is in line with the results of the high correlation between ROUGE 1 and the adequacy scores.

\begin{table}[t!]
    \centering
    \tabcolsep 3pt
    \begin{tabular}{lllllll}
        \hline
         & & Model & Accuracy & Precision & Recall & F1 \\
         \hline
         Task B & en & Ours & \textbf{0.977} & 0.735 & 0.926 & 0.820 \\
         &  & Majority &  0.944 & - & - & - \\
         & cs & Ours & \textbf{0.957} & 0.905 & 0.633 & 0.745 \\
         &  & Majority &  0.900 & - & - & - \\
         \hline
         Task C & en & Ours & \textbf{0.939} & 0.509 & 0.934 & 0.659 \\
         &  & Majority &  0.936 & - & - & - \\
         & cs & Ours &  \textbf{0.984} & 0.846 & 0.971 & 0.904 \\
         &  & Majority &  0.922 & - & - & - \\
         \hline
    \end{tabular}
    \caption{Overall results for our Task B and C submission.}
    \label{tab:task-bc-result}
\end{table}

\section{Conclusion}
In this paper, we have introduced the reference-free automatic minuting system and the document classification system using various surface-oriented features for the First Shared Task on Automatic Minuting (AutoMin-2021).
Given that reference minutes for Task A reflect diverse annotators' aspects, we aimed at building a reference-free pipeline rather than mimicking reference minutes. Thus, we adopted a module-based approach, which first splits a transcript into topic-specific blocks and subsequently summarizes those blocks with a pre-trained BART model fine-tuned on the SAMSum corpus. Motivated by the assumption that any coherent structure would improve readability, we further applied a technique of argument mining to the generated minutes, tailoring them in a well-structured and coherent way. Our approach achieved the best adequacy score among all submissions and showed competitive performance with the best system in terms of two human evaluation metrics: grammatical correctness and fluency. In addition, our approach demonstrated moderate ROUGE scores despite the reference-free approach.

To determine whether or not a minute is derived from the same meeting when either a transcript or another minute is given (Task B and C), we utilize multiple surface-based relevance scores. On top of those scores, we trained a conventional machine learning model, such as SVM, to bind them and made final decisions. Consequently, our approach demonstrated better results than the majority vote baselines. The results also suggest that some term overlapping-based features can be utilized as a useful evaluation metric to measure similarities between generated and reference minutes instead of ROUGE scores.

For future work, we intend to improve both grammatical correctness and fluency by paying more attention to the summarization module.
In addition, we will update our multilingual pipeline by referring to gold minutes with the best ROUGE 1 or task C scores as proxy variables of the adequacy scores. 

\bibliographystyle{IEEEtran}
\bibliography{custom,anthology}

\end{document}